\pdfoutput=1
\documentclass[11pt]{article}

\usepackage{acl}
\usepackage{hyperref}
\usepackage{times}
\usepackage{latexsym}

\usepackage{microtype}
\usepackage{multirow}
\usepackage{latexsym}
\usepackage{booktabs}
\usepackage{courier}
\usepackage{amsmath}
\usepackage{booktabs} % For better table formatting
\usepackage{siunitx} % For aligning numbers by decimal point
\usepackage{subfig}
\usepackage{placeins}
\usepackage{tikz}
\usepackage{hyperref}
\usepackage[T1]{fontenc}
\usepackage[utf8]{inputenc}

\usepackage{microtype}

\usepackage{inconsolata}

\title{DKE-Research at SemEval-2024 Task 2: Incorporating Data Augmentation with Generative Models and Biomedical Knowledge to Enhance Inference Robustness}

\author{Yuqi Wang\textsuperscript{1,3}, Zeqiang Wang\textsuperscript{4}, Wei Wang\textsuperscript{1}, Qi Chen\textsuperscript{1}, Kaizhu Huang\textsuperscript{2}, Anh Nguyen\textsuperscript{3}, Suparna De\textsuperscript{4} \\
	         \textsuperscript{1}Xi'an Jiaotong Liverpool University, China \hspace{2mm}
            \textsuperscript{2}Duke Kunshan University, China\\
          	\textsuperscript{3}University of Liverpool, United Kingdom \hspace{2mm}
      		 \textsuperscript{4}University of Surrey, United Kingdom\\
      	 	 \texttt{yuqi.wang17@student.xjtlu.edu.cn, \{wei.wang03,qi.chen02\}@xjtlu.edu.cn,}\\
      	 	 \texttt{kaizhu.huang@dukekunshan.edu.cn,}
      	 	   \texttt{anh.nguyen@liverpool.ac.uk}, \\
            \texttt{\{zeqiang.wang, s.de\}@surrey.ac.uk} }

\begin{document}
\maketitle
\begin{abstract}
Safe and reliable natural language inference is critical for extracting insights from clinical trial reports but poses challenges due to biases in large pre-trained language models. This paper presents a novel data augmentation technique to improve model robustness for biomedical natural language inference in clinical trials. By generating synthetic examples through semantic perturbations and domain-specific vocabulary replacement and adding a new task for numerical and quantitative reasoning, we introduce greater diversity and reduce shortcut learning. Our approach, combined with multi-task learning and the DeBERTa architecture, achieved significant performance gains on the NLI4CT 2024 benchmark compared to the original language models. Ablation studies validate the contribution of each augmentation method in improving robustness. Our best-performing model ranked 12th in terms of faithfulness and 8th in terms of consistency, respectively, out of the 32 participants.
\end{abstract}

\section{Introduction}

In the domain of clinical trial analysis, researchers and practitioners are overwhelmed with an ever-expanding corpus of clinical trial reports (CTRs). The current repository contains a vast number of documents and is rapidly growing, a trend that correlates with the increasing prevalence of cross-national, cross-ethnic, and multi-center clinical studies \cite{bastian2010seventy}. This growth necessitates a scalable approach to evaluate and interpret the massive amount of data in these reports \cite{goldberg2017evolution, li2020clinical}.

Recent advances in Natural Language Processing (NLP) offer promising avenues for the automated analysis of CTRs. Such analyses include medical evidence understanding \cite{nye2021understanding}, information retrieval \cite{wang2023zero}, causal relationship identification \cite{cai2017identification}, and the inference of underlying reasons for trial outcomes \cite{steinberg2023using}. Integrating natural language inference (NLI) with CTRs has the potential to revolutionize the large-scale, NLP-based examination of experimental medicine \cite{kim2018medical}.
Despite the progress in NLP, the application of large language models to this task presents several challenges, including susceptibility to shortcut learning, hallucination, and biases stemming from word distribution patterns within the training data \cite{huang2023survey}.

To address these issues, we propose a novel method that leverages generative language models, such as GPT-3.5\footnote{https://openai.com/chatgpt}, and biomedical domain knowledge graphs to enhance data diversity. Our approach introduces three types of data augmentation: numeric question-answering data generation, semantic perturbations, and domain-tailored lexical substitutions for the biomedical field. By combining these data augmentation techniques with multi-task learning and the DeBERTa 
 \cite{he2021deberta} architecture, we have achieved significant improvements in terms of faithfulness and consistency on the NLI4CT 2024 dataset. This paper outlines our approach, elaborates on the design of the perturbations and the multi-task learning process, and demonstrates the efficacy of our method through rigorous evaluation.

\section{Background}

In a crucial field like healthcare, where misinterpretations can have severe implications, NLI models must present precise predictions and reliable interpretations. This highlights the importance of accurate and trustworthy reasoning in these NLI models. 

SemEval 2024 Task 2 \cite{jullien-etal-2024-semeval} provides multi-sentence textual data consisting of patient case histories and medical reports. The objective of this task is to predict the logical relationship between the CTR and a given statement, including entailment and contradiction. The evaluation emphasizes prediction accuracy as well as the robustness to the controlled interventions, helping increase healthcare practitioners' trust in the system's predictions.

Enhancing the robustness of NLI models for healthcare can be strategically achieved using data augmentation techniques. Synthetic data generation via techniques like conditional text generation can expand training data diversity and volume to improve model generalization capabilities \cite{liu2020data, puri2020training, bayer2023data}. Meanwhile, multi-task learning with auxiliary objectives related to logical reasoning and explanation generation can enhance faithful reasoning abilities \cite{li2022faithfulness}. Useful domain knowledge can be captured by training language models on domain-specific medical textual datasets \cite{singhal2023large, tian2024opportunities}. Complementary data-centric methods can augment model architecture design to develop more capable, trustworthy, and clinical NLI systems.

\section{System overview}
\begin{figure*}[ht]
    \centering
    \includegraphics[width=\textwidth]{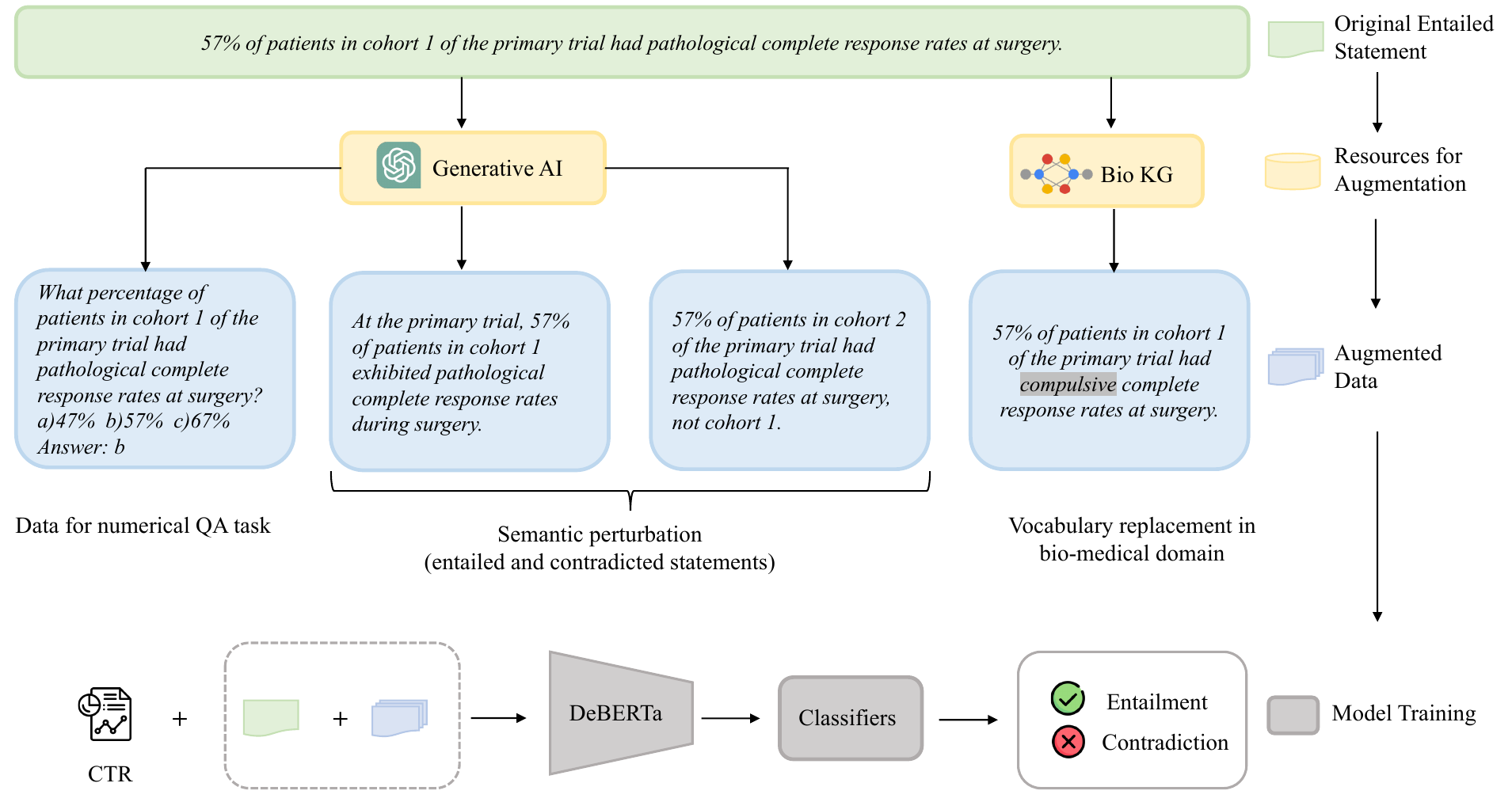}
    \caption{The overall demonstration of the proposed system. The upper part of the demonstration involves the application of data augmentation techniques to entailed statements extracted from the original NLI dataset, leveraging generative artificial intelligence (AI) and biomedical domain knowledge graphs. Specifically, we undertake the following procedures: 1) Transformation of statements into multiple-choice questions accompanied by corresponding answers; 2) Introduction of semantic perturbations to the original entailed statements; 3) Employing a statistical method to identify keywords within the original entailed statements, followed by their substitution with synonyms sourced from the biomedical knowledge graph. In the lower part of the demonstration, we incorporate the original entailed statements, augmented data, and CTRs as training data to develop a classifier based on the DeBERTa architecture. }
    \label{fig:demo}
\end{figure*}

In this section, we describe the proposed system to tackle the NLI problem and enhance the model's robustness against interventions spanning numerical, vocabulary, and semantic dimensions, as shown in Figure \ref{fig:demo}. 
\subsection{Data for Numerical Question Answering Task}
A major limitation of many language models lies in their tendency to learn linguistic patterns and features from large-scale textual data while lacking capabilities for numerical and quantitative reasoning \cite{geva2020injecting}. Such capabilities are crucial for analyzing relationships between CTRs and corresponding claims. Although BERT-based models pre-trained on NLI tasks, i.e. DeBERTa, can conduct general linguistic inference, they remain vulnerable to numerical perturbations in statements. 

Therefore, we propose to leverage GPT-3.5 to generate data tailored to the numerical question-answering task based on original entailed statements: The entailed statement, denoted as $x$, corresponding to a given CTR, is converted into a question $q$ that requires numerical reasoning. Subsequently, three candidate choices $c$ are enumerated, each accompanied by an answer $a$ extracted from the original statements. The loss function employed for this task is binary cross-entropy and is expressed as follows:
\begin{equation}\label{eq:loss1}
		\mathcal{L}_{NQA} = \left\{\begin{array}{lll}
			 -\log g\left(\text{CTR}, q, c; \theta_g\right)&& c = a\\ 
			
		   -\left[1 - \log g\left(\text{CTR}, q, c; \theta_g\right) \right]  &&c \ne a\\ 
		\end{array}\right.
\end{equation}
where  $g(\cdot)$ is the function to determine if the candidate choice is the correct answer and $\theta_g$ is the corresponding parameters for the DeBERTa backbone network and the additional classifier. 

This numerical question-answering task serves as an auxiliary task to enhance numerical reasoning abilities. The final loss function for the system combines the losses from this task and the main NLI task, i.e.
\begin{equation}
    \mathcal{L} = \mathcal{L}_{NLI} + \lambda\mathcal{L}_{NQA}
\end{equation}
where $\lambda$ is the hyper-parameter to be tuned in the validation phase.
\subsection{Semantic Perturbation}
We utilize GPT-3.5 to generate perturbed statements based on the original entailed input, obtaining both semantic-altering variants labeled as ``contradictions'' and semantic-preserving variants labeled as ``entailment''. Specifically, to produce contradictory versions, guiding keywords such as ``contradicted'' and ``minor changes'' are injected into the input prompt to slightly modify the original statement while altering the semantics to create a contradiction. Conversely, to generate entailed versions, guiding phrases such as ``paraphrase'' are included in the prompt to rephrase the statement extensively while retaining semantic equivalence. This controlled semantic perturbation of the input statement via guided text generation allows us to efficiently augment the dataset with both contradicting and entailing variants of the original input.
\subsection{Vocabulary Replacement}
When we analyze textual data in the clinical domain, we need to pay attention to the vocabulary because it contains many terms that are specific to this domain \cite{wang2018comparison}. However, most NLI models are pre-trained on data from general domains, and they are unaware of the meaning or relevance of these terms \cite{wang2023fusing}. To address this problem, we use a combination of biomedical knowledge graph embedding and statistical model, which can help us find the most important keyword to replace the term in the statement and generate the augmented data to improve the vocabulary alignment.
Specifically, given a statement $x$, consisting of $n$ words, i.e. $x =\{w_1, w_2, ..., w_n\}$ and the set of all the statements, denoted as $D$, we first remove all the stop-words and apply Term-Frequency-Inverse Document Frequency (TF-IDF) to identify the most important term in the statement, i.e.
\begin{equation}
    w^* =  \arg\max_{w_i \in x} \text{TF}(w_i, x) \times \text{IDF}(w_i, D)
\end{equation}
Subsequently, we locate a term in the biomedical embedding space that shares the same part-of-speech and has the highest similarity score with the chosen term, using it as the substitute, i.e.
\begin{equation}
    \hat{w}^* = \arg\max_{w \in V} \{sim(w^*, w)|\\
    \text{PoS}(w) = \text{PoS}(w^*)\}
\end{equation}
where $V$ is the biomedical term vocabulary and $\text{PoS}(\cdot)$ is the part-of-speech of a word. In this way, we can substitute $w^*$ in the original statement with $\hat{w}^*$ to generate a new adversarial sample to enhance the model robustness in the vocabulary aspect.
\section{Experimental setup}
\subsection{Dataset}
\begin{table}[htp]
\centering
\begin{tabular}{lllll|l}
\toprule
     & Ent. & Con. & Alt.  & Pres. & SUM   \\
     \midrule
Train & 850 & 850 & - & - & 1,700 \\
Val.  & 100  & 100  & 1,606 & 336   & 2,142 \\
Test & 250  & 250  & 4,136 & 864   & 5,500 \\
\bottomrule
\end{tabular}
\caption{Statistics of the validation and test set. ``Ent.'' and ``Con.'' stands for entailment and contradiction, while ``Alt.'' and ``Pres.'' stands for altering and preserving. }
\label{tab:data}
\end{table}
We conducted experiments on the NLI4CT 2024 dataset \cite{jullien-etal-2024-semeval}, generated by clinical domain experts and sourced from a large database for clinical studies\footnote{https://ClinicalTrials.gov}. The statistic of this dataset is summarized in Table \ref{tab:data}. The training data is the same as the NLI4CT 2023 dataset \cite{jullien-etal-2023-nli4ct} while there are perturbed samples in the validation and testing sets. 
\subsection{Metrics}
We first assessed the performance of the original statements without any perturbation and recorded the corresponding F1 score, precision, and recall. Then, we assessed the performance of the contrast set, consisting of interventions. Specifically, to evaluate the model's robustness to the semantic-preserving interventions, we used consistency as the metric, i.e.
\begin{equation}
\footnotesize
    \begin{aligned}
   Consistency = \frac{1}{N}\sum_{1}^{N} 1 - \left| f(x_i^\prime)-f(x_i) \right| \\ x_i^\prime\in C:\text{Label}(x_i) = \text{Label}(x_i^\prime)
    \end{aligned}
\end{equation}
Where $C$ is the contrast set, and $N$ is the number of the statements in the contrast set. $x_i^\prime$ is the perturbed statement for $x_i$ and $f(\cdot)$ computes the final prediction from the model. For the semantic-altering interventions, we evaluated the model using faithfulness, i.e.
\begin{equation}
\footnotesize
    \begin{aligned}
   Faithfulness = \frac{1}{N}\sum_{1}^{N}\left| f(x_i^\prime)-f(x_i) \right|\hspace{1cm}\\ x_i^\prime\in C:\text{Label}(x_i) \neq \text{Label}(x_i^\prime), \text{ and } f(x_i) = \text{Label}(x_i) 
    \end{aligned}
\end{equation}
\subsection{Implementation details}
\begin{table*}[ht]
\centering
\begin{tabular}{l|l}
\toprule
                     & Prompt                     \\ \midrule
\multirow{5}{*}{NQA} & \multirow{5}{*}{\begin{tabular}[c]{@{}l@{}}\textit{Please convert the statement to a multiple choice question that requires the numerical} \\ \textit{or quantitative reasoning, and each question has 3 choices, }\\ \textit{using the given template: }\textbackslash{}\textit{n}\\ \textit{Question:} \textit{{[}Question}{]} \textbackslash{}\textit{n }\textit{Choices: 1. }{[}\textit{Choice 1}{]}\textbackslash{}\textit{n} \textit{2.} {[}\textit{Choice 2}{]}\textbackslash{}\textit{n} \textit{3.} {[}\textit{Choice 3}{]}\textbackslash{}\textit{n}\\ \textit{Correct Answer: {[}Correct Answer{]}.} \\\end{tabular}} \\ 
                     &                            \\
                     &                            \\
                     &                            \\
                     &                            \\\midrule
SP.-Ent.             & \textit{Please rephrase the given statement:}            \\ \midrule
\multirow{2}{*}{SP.-Con.} & \multirow{2}{*}{\begin{tabular}[c]{@{}l@{}}\textit{Please generate a contradictory statement based on the given statement,}\\ \textit{with a minor change:}\end{tabular}} \\
                          &     \\                                                                                                                    
\bottomrule
\end{tabular}
\caption{Prompts for numerical question-answering data generation and semantic perturbation. NQA stands for numerical question answering. SP.-Con. and SP.-Ent. means semantic perturbation to generate statements labeled as contradiction and entailment, respectively.}
\label{tab:prompt}
\end{table*}
We downloaded DeBERTa models from the Huggingface repository\footnote{https://huggingface.co/} and implemented our proposed method based on Python 3.10 and Pytorch 2.1.1. During the model training, we used the Adam optimizer and set the learning rate to $5e-6$ with a batch size of 4, following the original work \cite{he2021deberta}. The maximum sequence length the model can take was set to $512$. The epoch number was set to $20$, and the early stopping based on the validation set was applied to avoid overfitting. The input format for the NLI task in this work is structured as follows: [CLS] + CTR + [SEP] + claim + [SEP]. In this structure, [CLS] serves as the initial token for classification in DeBERTa, and [SEP] acts as a separator token. For the vocabulary replacement, we used the bio-medical domain embedding from the work by \cite{zhang2019biowordvec}, which has been pre-trained over the MeSH knowledge graph\footnote{https://www.ncbi.nlm.nih.gov/mesh/}. For preprocessing, such as stop word filtering and part-of-speech tagging, we used the NLTK library\footnote{https://www.nltk.org/} in Python.
We include prompts for numerical question-answering data generation and semantic perturbation in Table \ref{tab:prompt}.

\section{Results}
\begin{table*}[htp]
\centering
\begin{tabular}{lccccc|ccccc}
\toprule
\multirow{2}{*}{\textbf{Method}} & \multicolumn{5}{c}{\textbf{Validation}}                                            & \multicolumn{5}{c}{\textbf{Test}}      
\\
\cline{2-6}
\cline{7-11}
                                & \textbf{F1}    & \textbf{Prec.} & \textbf{Rec.} & \textbf{Faith.} & \textbf{Con.}  & \textbf{F1}    & \textbf{Prec.} & \textbf{Rec.} & \textbf{Faith.} & \textbf{Con.}  \\
                                \midrule
\textbf{DeBERTa-l}                
& \textbf{81.82} & 90.00          & 75.00         & 73.81           & 71.48          & \textbf{77.25} & 80.80          & 73.99         & 67.13           & 71.06          \\
\textbf{+SP}                    & 81.77          & 83.00          & 80.58         & 85.42           & \textbf{75.16} & 75.52          & 72.80          & 78.45         & 78.24           & 74.01          \\
\textbf{\hspace{2mm}+VR}                   & 81.00          & 81.00          & 81.00         & 86.01           & 74.16          & 75.05          & 71.60          & 78.85         & 78.59           & 74.42          \\
\textbf{\hspace{4mm}+NQA}                    & 80.60          & 81.00          & 80.20         & \textbf{86.61}  & 74.91          & 74.09          & 69.20          & 79.72         & \textbf{79.98}  & \textbf{74.54} \\
\midrule
\textbf{DeBERTa-b}           & 70.87 & 73.00 & 68.87 & 49.40  & 60.02 & \textbf{62.53} & 60.40 & 64.81 & 57.75  & 59.33 \\
\textbf{+SP}                    & \textbf{71.84 }& 74.00 & 69.81 & 51.49  & 60.65 & 62.08 & 59.60 & 64.78 & 60.65  & 59.70 \\
\textbf{\hspace{2mm}+VR }                  & 70.59 & 72.00 & 69.23 & 52.38  & 60.71 & 62.21 & 59.60 & 65.07 & 60.76  & 59.72 \\
\textbf{\hspace{4mm}+NQA}                    & 70.30 & 71.00 & 69.61 & \textbf{52.98}  & \textbf{60.77} & 62.05 & 59.20 & 65.20 & \textbf{61.92 } & \textbf{59.89}\\
\bottomrule
\end{tabular}
\caption{Results on the development set and testing set for NLI4CT 2024 dataset. DeBERTa-l and DeBERTa-b are the large version and base version of the DeBERTa model, respectively. SP and VR stand for semantic perturbation and vocabulary replacement. The best results for F1 score on the control set, faithfulness, and consistency are highlighted.}
\label{tab:res}
\end{table*}
We conducted experiments with different-sized DeBERTa models, iteratively adding augmented data from three different interventions to the training set. As shown in Table \ref{tab:res}, incorporating all three types of augmented data greatly improved the average faithfulness and consistency scores. Specifically, we witnessed gains of 8.17\% on DeBERTa-l and 2.37\% on DeBERTa-b. This result also suggests that the augmented training data provided more benefit to the larger-sized DeBERTa model in terms of robustness. The additional augmented examples may have provided useful regularization, helping it generalize better on both the unaltered control and contrast datasets. Our best-performing model ranked 12th in terms of faithfulness and 8th in terms of consistency, respectively, out of the 32 participants.

From this iterative process, we can see that semantic perturbation with generative AI contributes mainly to the performance gain for both NLI models. Compared with this, vocabulary replacement in the biomedical domain has only a minor effect. This may suggest that vocabulary replacement in our work may be relatively less effective in this case because it only swaps out individual words, while semantic perturbation modifies the whole statement. Hence, semantic perturbation provides more meaningful variations to augment the training data. 

While the augmented data improved the robustness to interventions, we noticed a slight performance drop in the control set. For example, the F1 score on the control set decreased by 3.16\% for DeBERTa-l and 0.48\% for DeBERTa-b after adding all the augmented data.
This performance decline indicates there may have been a small trade-off between improving robustness to interventions and maintaining strong performance on the original data. One of the reasons accounting for this could be that the generative AI may generate noisy or irrelevant data. For example, in numerical question answering data generation, if the original entailed statement discusses an assumption about a 50-year-old patient not mentioned in the CTR, the generative model may create an unrelated question about the patient’s age that cannot be inferred from the given information. Another example involves vocabulary replacement: we observed that there exist some cases where even two words having very similar embeddings in the biomedical domain knowledge graph embedding space may not be very closely related in the context of the current statement.
Including these illogical examples in the augmented training data could mislead the original DeBERTa model, resulting in worse performance on the unaltered control set.
\section{Conclusion}
In this work, we proposed a data augmentation approach to enhance the robustness of natural language inference models for clinical trial report analysis. Our method leverages generative AI and biomedical knowledge graphs to augment training data along three dimensions: numerical reasoning, semantic perturbations, and domain-tailored lexical substitutions. Experiments on the NLI4CT 2024 dataset demonstrate that our approach effectively improves model faithfulness and consistency against controlled interventions, with significant gains against the DeBERTa baselines. 

However, we observed a slight performance drop on the unaltered test set, indicating a trade-off between robustness to perturbations and maintaining strong performance on original data. Future work will focus on: 1) generating higher-quality augmented examples using numerical question-answering data generation to minimize or avoid performance drop; 2) validating the perturbed samples to help remove noisy or irrelevant examples \cite{wang2023generating}; 3) incorporating external structured knowledge via pre-training on knowledge graphs and not just lexical substitution, which can provide more contextual domain information.

\section{Acknowledgments}
We would like to thank all the anonymous reviewers for their valuable feedback. We would like to acknowledge the financial support provided by the Postgraduate Research Scholarship (PGRS) (contract number PGRS-20-06-013) at Xi’an Jiaotong-Liverpool University. Additionally, this research has received partial funding from the Jiangsu Science and Technology Programme (contract number BK20221260) and the Research Development Fund (contract number RDF-22-01-132) at Xi’an Jiaotong-Liverpool University.

\end{document}